\documentclass[letterpaper, 10 pt, conference]{ieeeconf}  

\IEEEoverridecommandlockouts                              

\usepackage{graphics} 
\usepackage{epsfig} 
\usepackage{mathptmx} 
\usepackage{times} 
\usepackage{amsmath} 
\usepackage{amssymb}  
\usepackage{booktabs}
\usepackage{tabularx}
\usepackage{multicol, blindtext}
\usepackage{comment}

\usepackage{lipsum}
\usepackage{color}
\usepackage[colorinlistoftodos]{todonotes}

\usepackage{algorithm}
\usepackage{algpseudocode}
\usepackage{pifont}

\usepackage{mathtools}
\usepackage{bm}
\usepackage{mathrsfs}
\usepackage{eucal} 

\usepackage{graphicx}
\usepackage{epstopdf}
\usepackage[font=footnotesize]{caption}
\usepackage{subcaption}
\usepackage{wrapfig}
\usepackage{tikz}
\usepackage{tabularx}
\usepackage{placeins}
\usepackage{dblfloatfix}

\usepackage[hang,flushmargin]{footmisc}   

\usepackage{float}
\usepackage{placeins}

\newcommand*\rot{\rotatebox{70}}

\title{\LARGE \bf
Analyzing Knowledge Transfer in Deep Q-Networks for Autonomously Handling Multiple Intersections
}

\author{David Isele$^{1}$, Akansel Cosgun$^{2}$ and Kikuo Fujimura$^{2}$
\thanks{$^{1}$David Isele is with The University of Pennsylvania, Philadelphia, PA, USA}
\thanks{$^{2}$Akansel Cosgun and Kikuo Fujimura are with the Honda Research Institute, Mountain View, CA, USA}
}

\begin{document}

\maketitle
\thispagestyle{empty}
\pagestyle{empty}

\begin{abstract}

We analyze how the knowledge to autonomously handle one type of intersection, represented as a Deep Q-Network, translates to other types of intersections (tasks). We view intersection handling as a deep reinforcement learning problem, which approximates the state action Q function as a deep neural network. Using a traffic simulator, we show that directly copying a network trained for one type of intersection to another type of intersection decreases the success rate. We also show that when a network that is pre-trained on Task A and then is fine-tuned on a Task B, the resulting network not only performs better on the Task B than an network exclusively trained on Task A, but also retained knowledge on the Task A. Finally, we examine a lifelong learning setting, where we train a single network on five different types of intersections sequentially and show that the resulting network exhibited catastrophic forgetting of knowledge on previous tasks. This result suggests a need for a long-term memory component to preserve knowledge.

\end{abstract}

\vspace{-0.2cm}

\section{Introduction}

Car companies has been increasing their R\&D spending on Automated Driving (AD) technology in recent years, for good cause: AD promises to greatly reduce accident-related fatalities and increase productivity of the society as a whole. Although AD technology has made important strides over the last couple of years, current technology is still not ready for large scale roll-out. Urban environments especially pose significant challenges for AD, due to the unpredictable nature of pedestrians and vehicles in city traffic. Handling intersections safely and efficiently is one of the most challenging problems for Urban AD.

Rule-based methods provide a predictable method to handle intersections. However, rule-based intersection handling approaches don't scale well because it becomes increasingly harder to design hand-crafted rules as scene complexity increases. Moreover, the algorithm designer has to come up with hand-crafted rules and parameters for different types of intersections. By different intersection types, we mean single or multi-lane right, left turns and forward passing.

Our goal for this research is to explore a machine learning based method that generalizes to various types of intersections. Machine learning, and particularly deep learning is a growing field that had tremendous impact on applications such as computer vision, speech recognition and language translation and it is increasingly being used for decision making. We model the AD vehicle as a learning agent, which learns from positive (successful passing) and negative experiences (collisions) in a reinforcement learning framework.

\begin{figure}[t]
    \centering
    \begin{subfigure}[b]{1.5in}
    	\includegraphics[width=.95\textwidth]{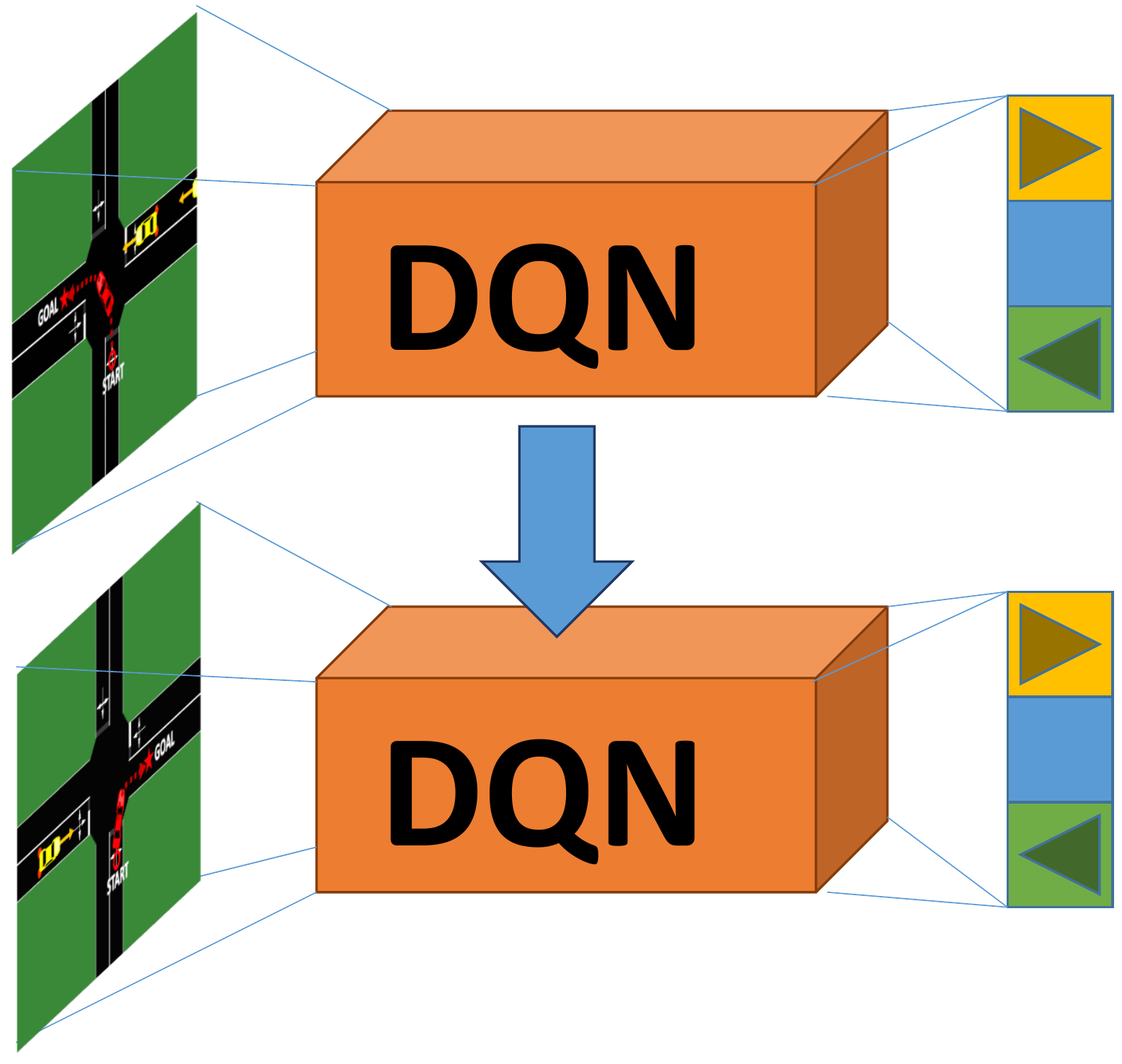} 
        \caption{Direct Copy}        
        \label{fig:transfer0}
    \end{subfigure}
    \begin{subfigure}[b]{1.5in}
    	\includegraphics[width=.95\textwidth]{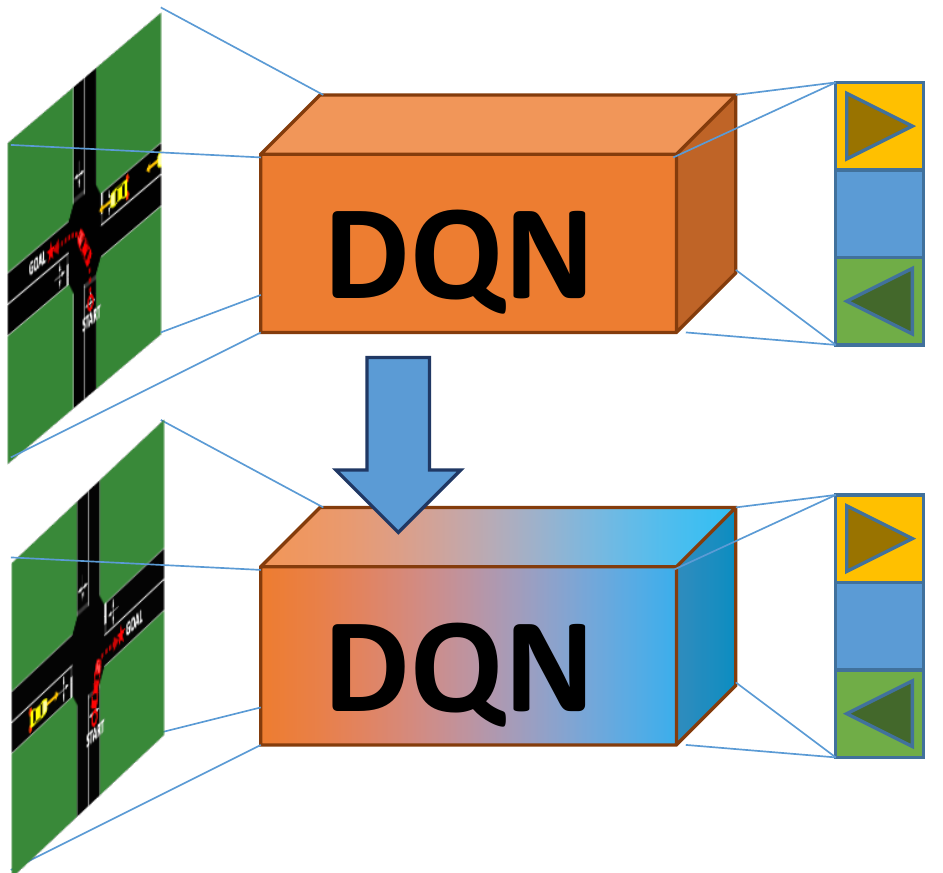}
        \caption{Fine Tuning}
        \label{fig:transfer1}        
    \end{subfigure}   
    \par\medskip
    \begin{subfigure}[b]{1.5in}
    
    	\includegraphics[width=.95\textwidth]{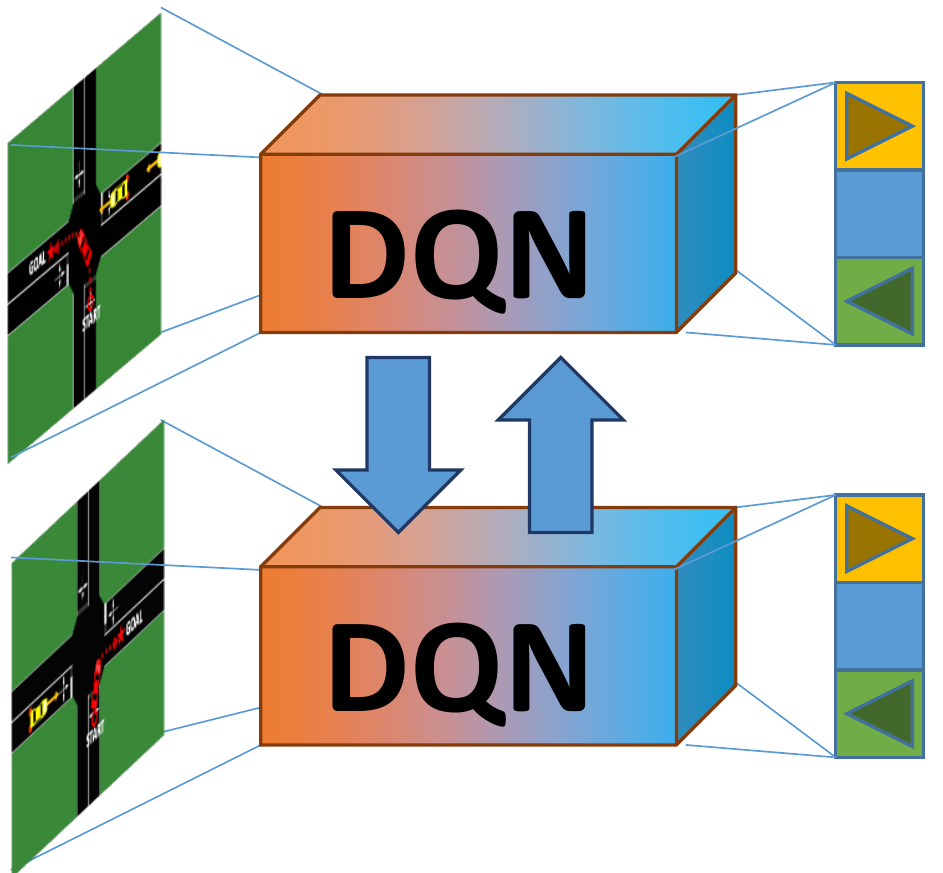}
        \caption{Reverse Transfer}
        \label{fig:transfer2}
    \end{subfigure}    
    \begin{subfigure}[b]{1.5in}
    \hspace{0.6cm}
       \includegraphics[width=.7\textwidth]{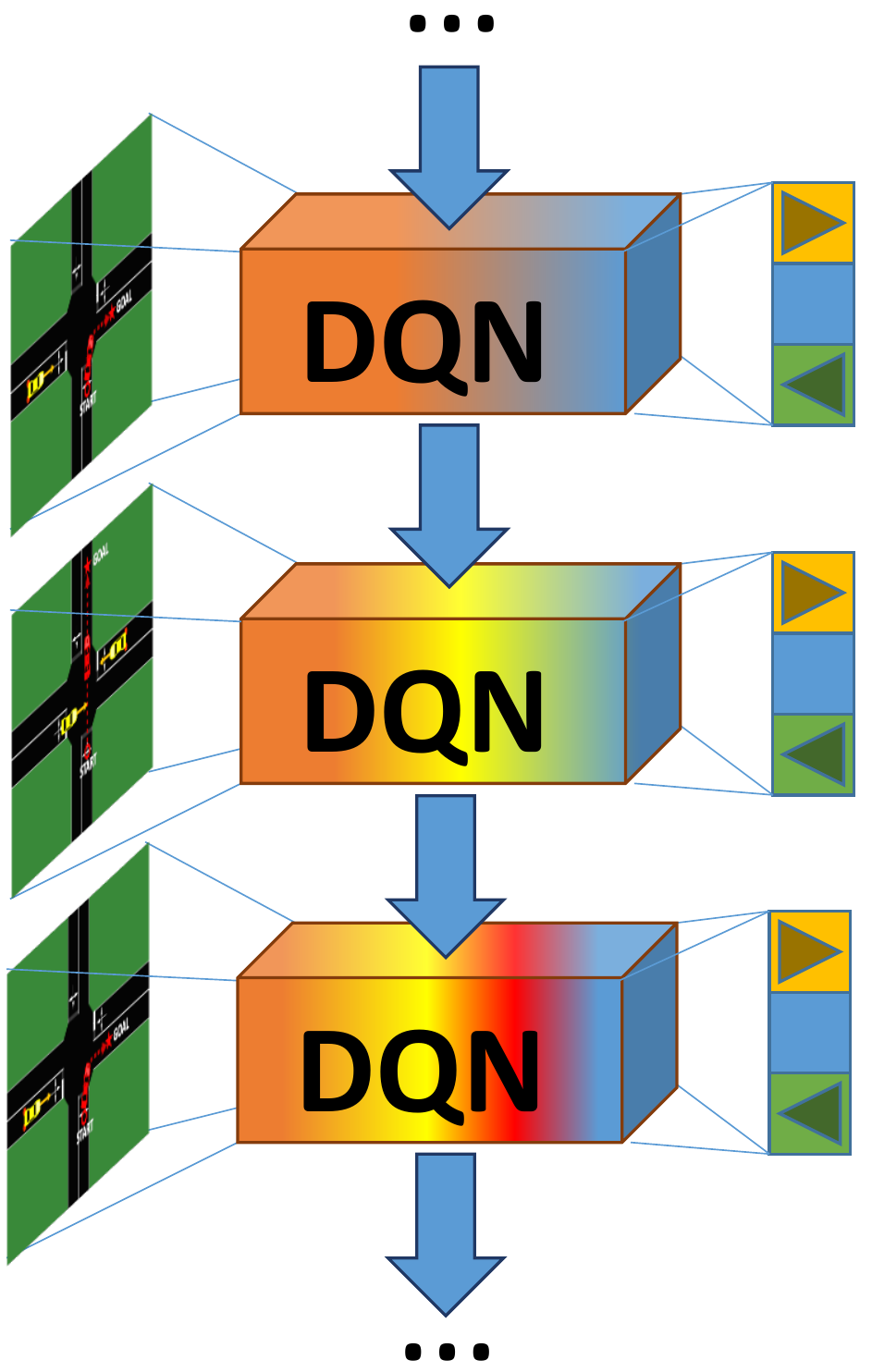}
        \caption{Lifelong Learning}
        \label{fig:transfer3}        
    \end{subfigure}
    \vspace{0.2cm}
        \caption{We investigate 4 types of knowledge transfer between different types of intersections. The knowledge on how to handle an intersection is represented as a Deep Q-Network (DQN), which is trained from simulation data. At every intersection the DQN makes a decision to either wait at the intersection or go. We analyze a) directly copying a deep network to a new intersection b) fine tuning a previously trained deep network on a new intersection, c) whether fine tuning destroys old intersection knowledge in reverse transfer and d) lifelong learning of multiple intersections with a single deep neural network.}\label{fig:transfers}
        \vspace{-0.35cm}
\end{figure}

In this paper we focus on how the knowledge for one type of intersection, represented as a Deep Q-Network (DQN), translates to other types of intersections (tasks). First we look at \textbf{direct copy}: how well a network trained for Task A performs in Task B. Second, we analyze how the performance of a network initialized from Task A and \textbf{fine tuned} in Task B compares to a randomly initialized network exclusively trained on Task B. Third, we investigate \textbf{reverse transfer}: if a network pre-trained for Task A and fine-tuned to Task B, preserves knowledge for Task A. Finally, we explore training a network for five tasks sequentially as a \textbf{lifelong learning} scenario.

This paper is organized as follows. After providing a brief literature survey in Section \ref{sec:related_works}, we present the problem formulation as a DQN in Section \ref{sec:intersection_handling}, before examining various knowledge sharing strategies in Section \ref{sec:knowledge}. After explaining the experimental setup in Section \ref{sec:experiments}, then present our results in Section \ref{sec:results} before concluding in Section \ref{sec:conclusion}.

\section{Related Work}
\label{sec:related_works}

Recently there has been an increased interest in using machine learning techniques to control autonomous vehicles. In imitation learning, the policy is learned from a human driver \cite{bojarski2016end}.
Online planners based on partially observable Monte Carlo Planning (POMCP) have been shown to handle intersections \cite{bouton2017belief} if the existence of an accurate generative model is available, and Markov Decision Processes (MDP) have been used offline to address the intersection problem \cite{brechtel2014probabilistic,song2016intention}.
Additionally, machine learning has been used to optimize comfort from a set of safe trajectories \cite{shalev2016safe}.

Machine learning has greatly benefited from training on large amounts of data. This helps a system learn general representations and prevents over fitting based on incidental correlations in the sampled data. In the absence of huge datasets, training on multiple related tasks can give similar improvement gains \cite{Caruana1997}. A large breadth of research has investigated transferring knowledge from one system to another both in machine learning in general \cite{Pan2010a}, and reinforcement learning specifically \cite{taylor2009transfer}. 

The training time and sample complexity of deep networks make transfer methods particularly appealing \cite{Razavian2014}, and has prompted in depth investigation to help understand its behavior \cite{Yosinski2014}. Recent work in deep reinforcement learning has looked at combining networks from different tasks to share information  \cite{rusu2016progressive,yin2017knowledge}. And efforts have been made to enable a unified framework for learning multiple tasks through changes in architecture design \cite{srivastava2013compete} and modified objective functions \cite{kirkpatrick2016overcoming} to address known problems like catastrophic forgetting \cite{goodfellow2013empirical}.



\section{Intersection Handling using Deep Q-Networks}
\label{sec:intersection_handling}

We view intersection handling as a reinforcement learning problem, and use a Deep Q-Network (DQN) to learn the state action value Q-function. We assume the AD vehicle is at the intersection, the path is known to it, and the network is tasked with choosing between two actions: wait or go, for every time step. Once the agent decides to go, it follows an intelligent driver model for keeping distance with the vehicles in front.

\subsection{Reinforcement Learning}

In reinforcement learning, an agent in state $s$ takes an action $a$ according to the policy $\pi_\theta$ parameterized by $\theta$. The agent transitions to the state $s'$
, and receives a reward $r$. This collection is defined as an experience  $e = (s,a,r,s')$. 

This is typically formulated as a Markov Decision Process (MDP) $\langle \mathcal{S}, \mathcal{A}, P, \mathcal{R}, \gamma  \rangle$, where $\mathcal{S}$ is the set of states,  $\mathcal{A}$ is the set of actions that the agent may execute, \mbox{$P: \mathcal{S} \times \mathcal{A} \rightarrow \mathcal{S}$} is the state transition function, $\mathcal{R}: \mathcal{S} \times \mathcal{A} \times \mathcal{S} \rightarrow \mathbb{R} $ is the reward function, and $\gamma \in (0,1]$ is a discount factor that adds preference for earlier rewards and provides stability in the case of infinite time horizons.
MDPs follow the Markov assumption that the probability of transitioning to a new state given the current state and action is independent of all previous states and actions $p(s_{t+1}|s_{t}, a_{t}, \dots, s_{0}, a_{0}) = p(s_{t+1}|s_{t}, a_{t})$. 

The goal at any time step $t$ is to maximize the future discounted return $R_t = \sum_{k=t}^T \gamma^{k-t} r_{k}$.
In order to optimize the expected return we use Q-learning \cite{watkins1992q}. 

\subsection{Q-learning}
Q-learning defines an optimal action-value function $Q^*(s,a)$ as the maximum expected return that is achievable following any policy given a state $s$ and action $a$, $Q^*(s,a) = max_\pi \mathbb{E}[R_t|s_t = s, a_t = a, \pi]$. 

This follows the dynamic programming properties of the Bellman equation, which state that if the values $Q^*(s',a')$ are known for all $a'$ then the optimal strategy is to select $a'$ that maximizes the expected value of $r+\gamma Q^*(s',a')$:
\begin{eqnarray}
Q^*(s,a) = \mathbb{E} [r + \gamma \max_{a'} Q^*(s',a')|s,a]  \enspace.
\end{eqnarray}

In Deep Q-learning \cite{mnih2013playing}, the optimal value function is approximated with a neural network $Q^*(s,a)\approx Q(s,a;\theta)$. The parameters $\theta$ are learned by using the Bellman equation as an iterative update $Q_{i+1}(s,a) = \mathbb{E} [r + \gamma \max_{a'} Q_i(s',a')|s,a]$ and minimizing the error between the expected return and the state-action value predicted by the network. This gives the loss for an individual experience in a deep Q-network (DQN)   
\begin{eqnarray}
\mathcal{L}(e_i,\theta) = \bigg( r_i + \gamma \max_{a_i'}Q(s_i',a_i';\theta) - Q(s_i,a_i;\theta)\bigg)^2    \enspace .
\end{eqnarray}

In practice, $Q(s_{t+1},a_{t+1};\theta)$ is a poor estimate early on, which can make learning slow since many updates are required to propagate the reward to the appropriate preceding states and actions. One way to make learning more efficient is to use $n$-step return\cite{peng1996incremental} $\mathbb{E}[R_t|s_t=s,a] \approx r_t + \gamma r_{t+1} + \dots + \gamma^{n-1} r_{t+n-1} + \gamma^{n} \max_{a_{t+n}}Q(s_{t+n},a_{t+n};\theta)$.

During learning, an $\epsilon$-greedy policy is followed by selecting a random action with probability $\epsilon$ to promote exploration and otherwise greedily selecting the best action $max_a Q(s,a;\theta)$ according to the current network. 
In order to improve the effectiveness of the random exploration we make use of dynamic frame skipping.
Frequently the same repeated actions is required over several time steps. It was recently shown that allowing an agent to select actions over extended time periods improves the learning time of an agent \cite{srinivas2017dynamic}. For example, rather than having to explore through trial and error and build up over a series of learning steps that eight time steps is the appropriate amount of time an agent should wait for a car to pass, the agent need only discover that a "wait eight steps" action is appropriate. Dynamic frame skipping can viewed as a simplified version of options \cite{sutton1998reinforcement} which is recently starting to be explored by the Deep RL community.
\cite{jaderberg2016reinforcement,tessler2016deep,kulkarni2016hierarchical}.

\begin{figure*}[t]
    \centering
    \vspace{8pt}
    \begin{subfigure}[b]{1.33in}
    	\includegraphics[height=.75\textwidth]{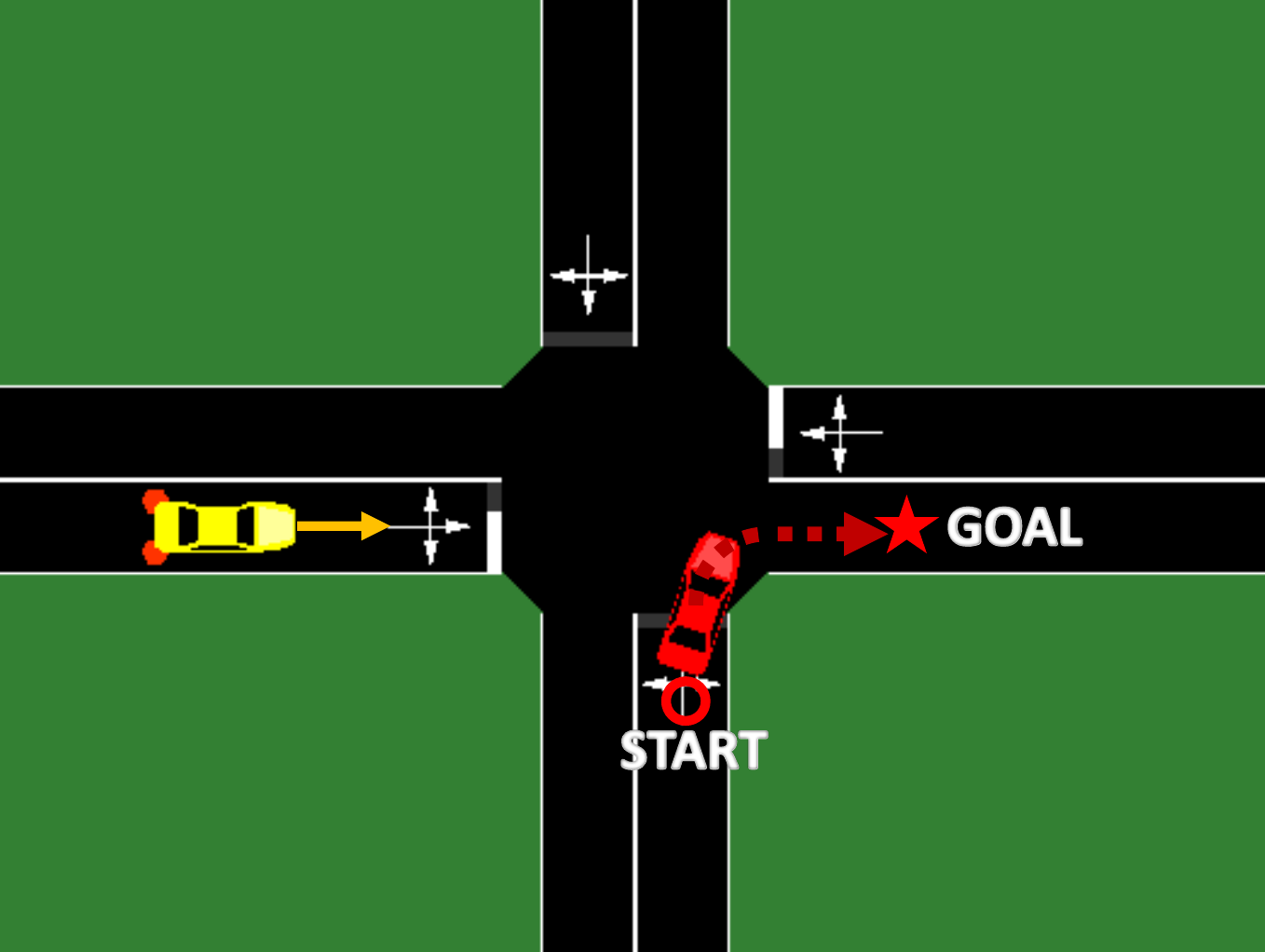} 
        \caption{\emph{Right}}
        \label{fig:scenarios_right}
    \end{subfigure}
    \begin{subfigure}[b]{1.33in}
    	\includegraphics[height=.75\textwidth]{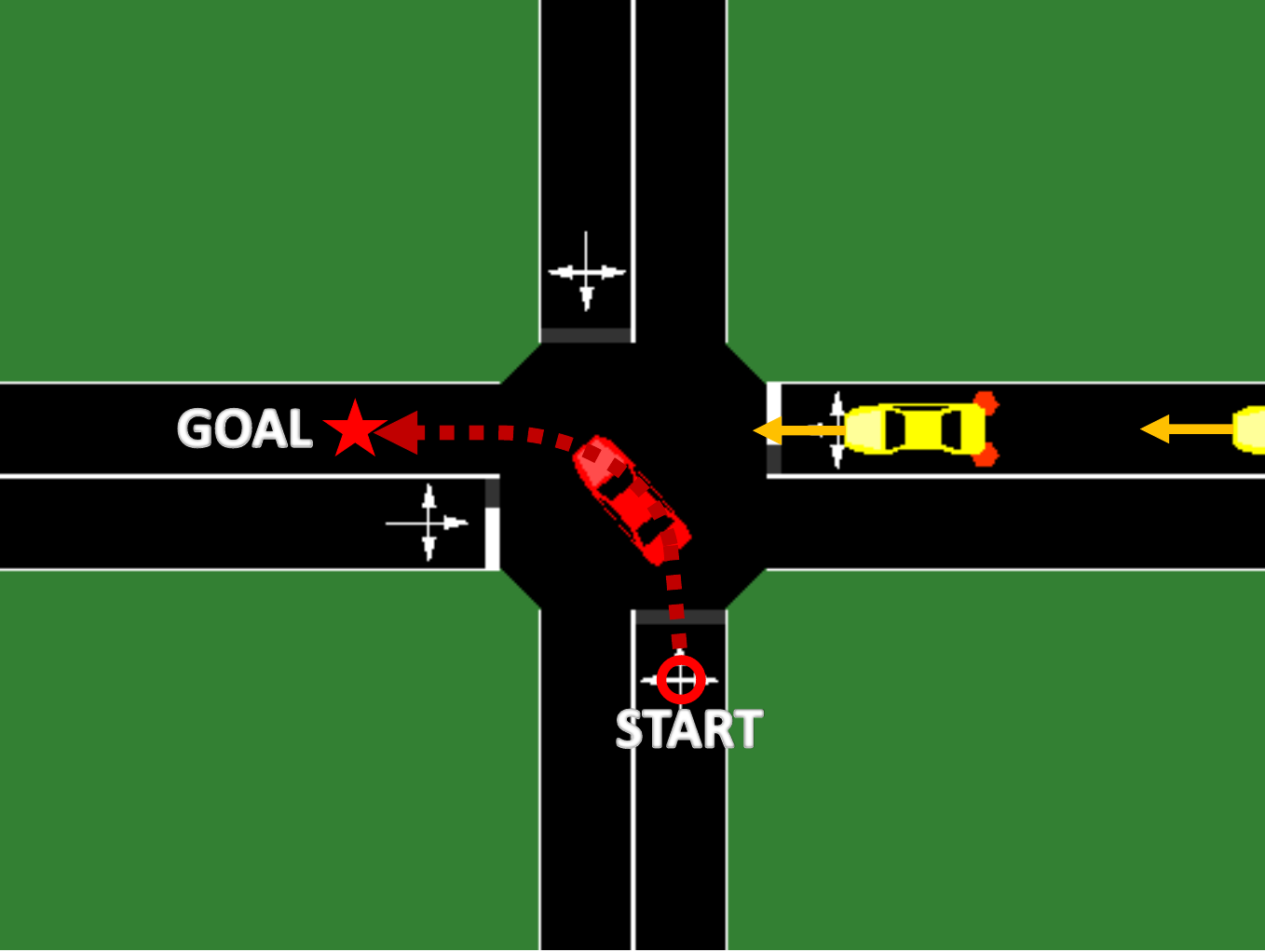}
        \caption{\emph{Left}}
        \label{fig:scenarios_left}
    \end{subfigure}
    \begin{subfigure}[b]{1.33in}
    	\includegraphics[height=.75\textwidth]{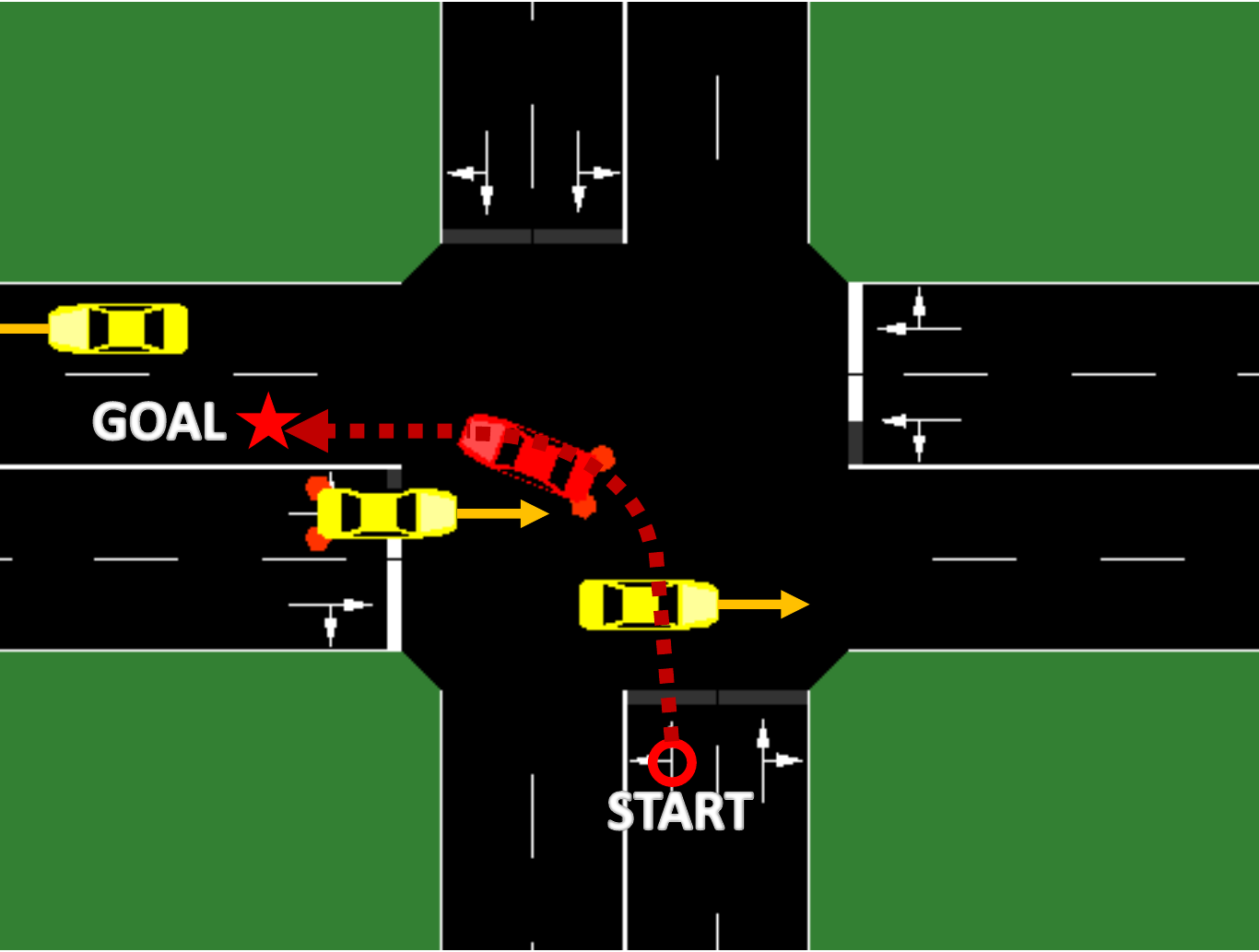}
        \caption{\emph{Left2}}
        \label{fig:scenarios_left2}
    \end{subfigure}
    \hspace{-4pt}
    \begin{subfigure}[b]{1.33in} 
       \includegraphics[height=.75\textwidth]{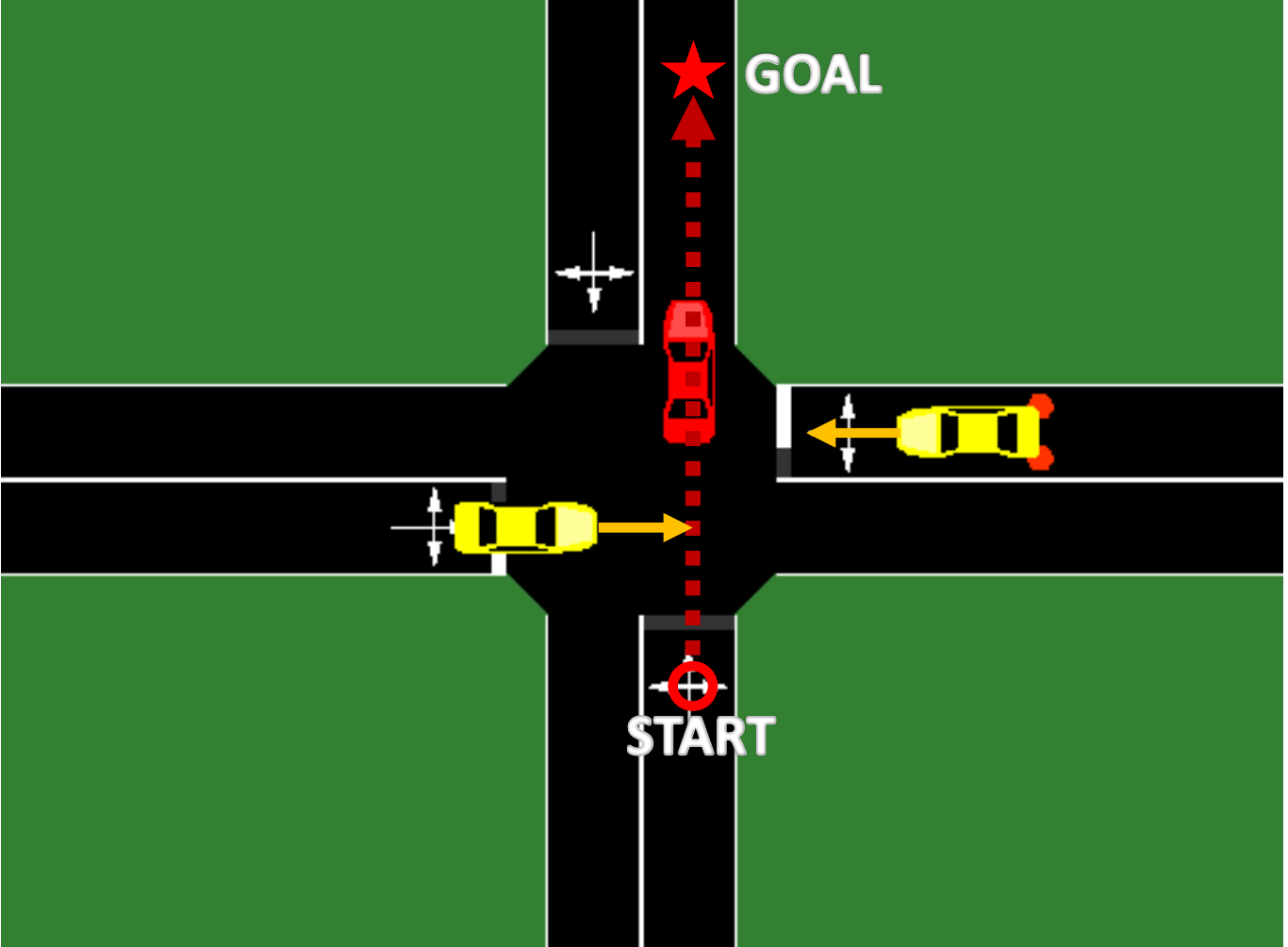}
        \caption{\emph{Forward}}
        \label{fig:scenarios_forward}
    \end{subfigure}
    \hspace{1pt}
    \begin{subfigure}[b]{1.33in}
    	\includegraphics[height=.75\textwidth]{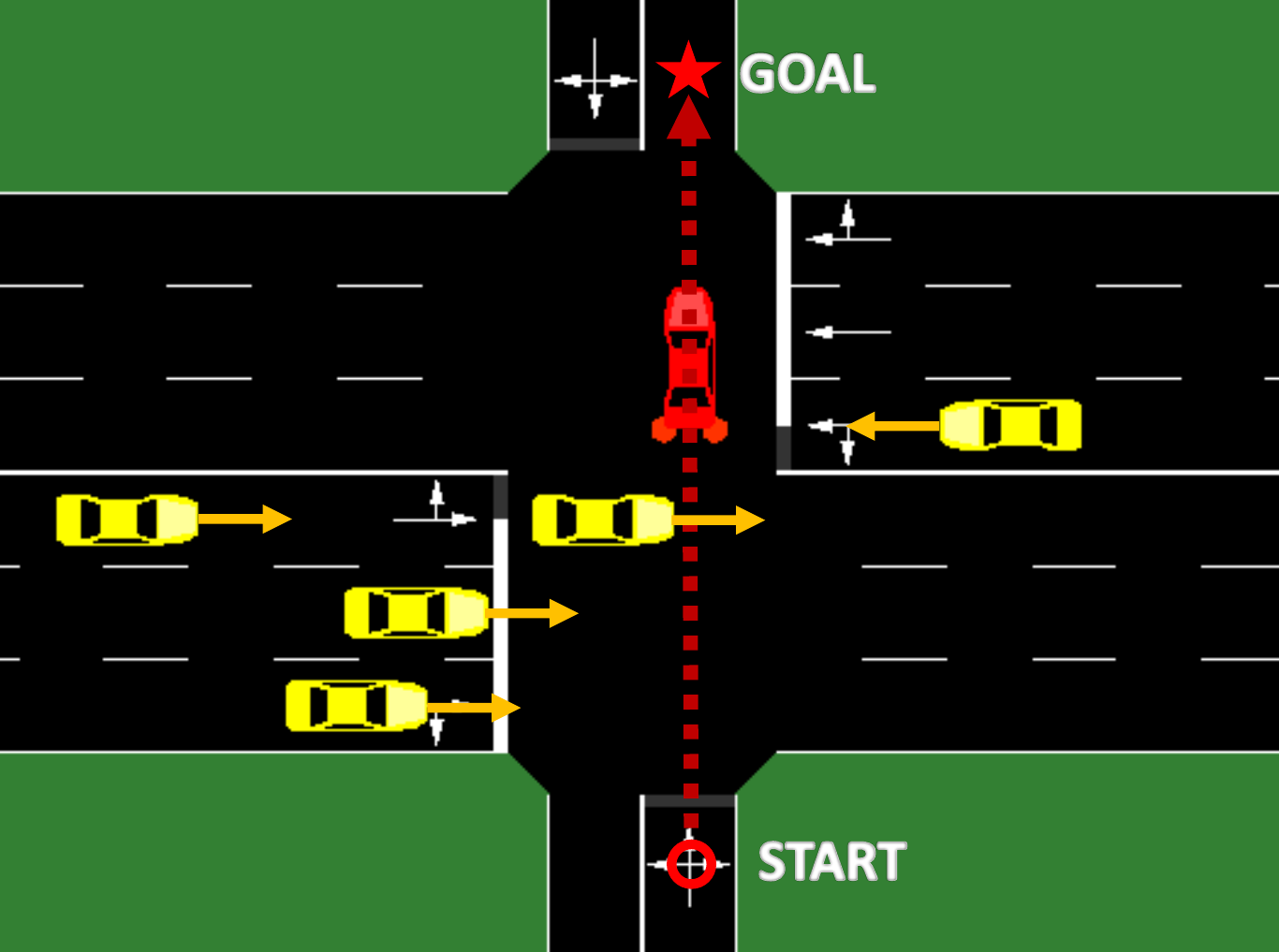}
        \caption{\emph{Challenge}}
        \label{fig:scenarios_challenge}
    \end{subfigure}
        \caption{Visualizations of different intersection scenarios.}\label{fig:scenarios} 
        \vspace{-10pt}
\end{figure*}

\subsection{Deep Neural Network setup}

The DQN uses a convolutional neural network with two convolution layers, and one fully connected layer. The first convolutional layer has $32$ $6 \times 6$ filters with stride two, the second convolution layer has $64$ $3 \times 3$ filters with stride 2. The fully connected layer has 100 nodes. All layers use leaky ReLU activation functions \cite{maas2013rectifier}. The final linear output layer has five outputs: a single \emph{go} action, and a \emph{wait} action at four time scales (1, 2, 4, and 8 time steps). The network is optimized using the RMSProp algorithm \cite{tieleman2012lecture}. 

Our experience replay buffers have an allotment of $1,000$ experiences. At each learning iteration we samples a batch of 60 experiences. 
Since the experience replay buffer imposes off-policy learning, we are able to calculate the return for each state-action pair in the trajectory prior to adding each step into the replay buffer. This allows us to train directly on the n-step return and forgo the added complexity of using target networks \cite{mnih2015human}.

The state space of the DQN is represented as a $18 \times 26$ grid in global coordinates. The epsilon governing random exploration was $0.05$. For the reward we used $+1$ for successfully navigating the intersection, $-1$ for a collision, and $-0.01$ step cost. 

\section{Knowledge Transfer}
\label{sec:knowledge}

We are interested in sharing knowledge between different driving tasks. By sharing knowledge from different tasks we can reduce learning time and create more general and capable systems. Ideally knowledge sharing can be extended to involve a system that continues to learn after it has been deployed \cite{Thrun1996} and can enable a system to accurately predict appropriate behavior in novel situations \cite{isele2016task}. We examine the behavior of various knowledge sharing strategies in the autonomous driving domain.

\subsection{Direct copy} 
To demonstrate the extent of transfer and show the difference between tasks, we train a network on a single source task for 25,000 iterations. The \emph{unmodified} network is then evaluated on every other task. We repeat this process, using each different task as a source task. 

\subsection{Fine tuning}
Starting with a network trained for 10,000 iterations on a \emph{source} task, we then fine tune a network for an additional 25,000 iterations on second \emph{target} task. We use 10,000 iterations because it demonstrates substantial learning, but is suboptimal in order to emphasize the possible benefits gained from transfer. Fine tuning demonstrates the \emph{jumpstart} and \emph{asymptotic performance} as described by Taylor and Stone\cite{taylor2009transfer}.

\subsection{Reverse transfer}

After a network has been fine tuned on the target task, we evaluate the performance of that network on the source task. If training on a later task improves the performance of an earlier task this is known as reverse transfer. It is known that neural networks often \emph{forget} earlier tasks in what is called catastrophic forgetting \cite{mccloskey1989catastrophic,ratcliff1990connectionist,goodfellow2013empirical}.  

In the case of forgetting, \emph{retention} describes the amount of previous knowledge retained by the network after training on a new task. This value is difficult to define formally since it must exclude any relevant knowledge for source tasks obtained from training on the target task, and additionally retention might include aspects that are not quantifiable such of weight configurations in the network. For example a network might exhibit catastrophic forgetting but in fact have retained a weight configuration that greatly reduces the training time needed to retrain the source task. Because of the difficulty of defining retention we define the \emph{empirical retention} as the difference between the direct copy and fine tuned direct copy of the same network.

\subsection{Lifelong Learning}

Lifelong learning is the process of learning multiple tasks sequentially where the goal is to optimize the performance on every task \cite{Thrun1996,Ruvolo2013}. The combination of information from all previous tasks can be used to jumpstart learning a new task. In a reciprocal fashion, learning a new task can potentially refine existing knowledge for previous tasks. By having a single system that handles all tasks, the system is able to handle a broader set of problems and will likely generalize better to new problems. 

We examine how a deep Q-network performs when learning a sequence of tasks. The order in which tasks are encountered does impact learning, and several groups have investigated the effects of ordering \cite{Bengio2009,ruvolo2013active}. For our experiments we use a task ordering that demonstrates forgetting and hold it fixed for all experiments.

We are interested in how each tasks performance changes over time. We test at regular intervals with testing run as a separate procedure that does not have an impact on the replay buffer or learning process of the network.

\section{Experimental Setup}
\label{sec:experiments}

Experiments were run using the Sumo simulator \cite{sumo}, which is an open source traffic simulation package. This package allows users to model road networks, road signs, traffic lights, a variety of vehicles (including public transportation), and pedestrians to simulate traffic conditions in different types of scenarios. Importantly for the purpose of testing and evaluation of autonomous vehicle systems, Sumo provides tools that facilitate online interaction and vehicle control. For any traffic scenario, users can have control over a vehicle's position, velocity, acceleration, steering direction and can simulate motion using basic kinematics models. Traffic scenarios like multi-lane intersections can be setup by defining the road network (lanes and intersections) along with specifications that control traffic conditions. To simulate traffic, users have control over the types of vehicles, road paths, vehicle density, and departure times. Traffic cars follow IDM to control their motion. In Sumo, randomness is simulated by varying the speed distribution of the vehicles and by using parameters that control driver imperfection (based on the Krauss stochastic driving model \cite{krauss1998sumo}). The simulator runs based on a predefined time interval which controls the length of every step.

We ran experiments using five different intersection scenarios: \emph{Right}, \emph{Left}, \emph{Left2}, \emph{Forward} and a \emph{Challenge}. Each of these scenarios is depicted in Figure \ref{fig:scenarios}.
The \emph{Right} scenario involves making a right turn, the \emph{Forward} scenario involves crossing the intersection, the \emph{Left} scenario involves making a left turn, the \emph{Left2} scenario involves making a left turn across two lanes, and the \emph{Challenge} scenario involves crossing a six lane intersection. 

The Sumo traffic simulator is configured so that each lane has a 45 miles per hour (20 m/s) max speed. The car begins from a stopped position. Each time step is equal to 0.2 seconds. The max number of steps per trial is capped 100 steps which is equivalent to 20 seconds. The traffic density is set by the probability that a vehicle will be emitted randomly per second. We use depart 
probability of 0.2 for \emph{each lane} for all tasks.

Navigating intersections involves multiple conflicting objectives. 
We evaluate four metrics in order to collect our statistics. The metrics are as follows:
\begin{itemize}
\item \textbf{Percentage of successes:} the percentage of the runs the car successfully reached the goal. This metric takes into both collisions and time-outs.
\item \textbf{Percentage of collisions:} a measure of the safety of the method. 
\item \textbf{Average time:} how long it takes a successful trial to run to completion.
\item \textbf{Average braking time:} the amount of time other cars in the simulator are braking, this can be seen as a measure of how disruptive the autonomous car is to traffic.
\end{itemize}
While there are multiple metrics, we focus on the percentage of success, which is the metric used in all our plots. All state representations ignores occlusion, assuming all cars are always visible.

\begin{figure}[t]
\centering
\vspace{-10pt}
\hspace{-10pt}
  \includegraphics[width=.4\textwidth]{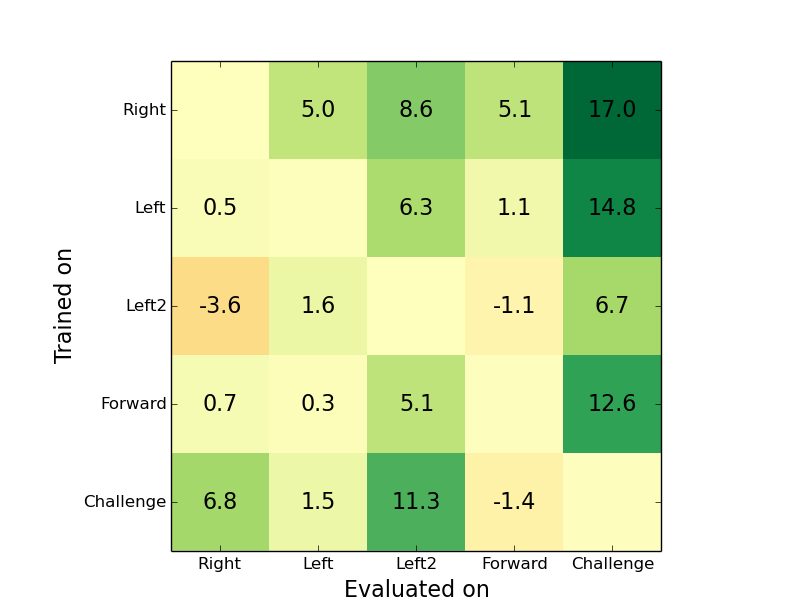}
  \caption{Retention. The table examines the performance of a network on the source task after training on the target task. We subtract the baseline performance of a network trained exclusively on the target task. This shows how much of the source task training exists after training on the target task.}
  \label{fig:reverse}\label{fig:retention}
\end{figure}

\begin{table}[htp]
\caption{Direct Copy Performance}
\begin{center}
\begin{tabular}{ccccccc}
\toprule 
    Task & Metric & \multicolumn{5}{c}{Training Method \vspace{-0.2cm}}\\    
    \cr & & \rot{\emph{Right}} & \rot{\emph{Left}} & \rot{\emph{Left2}} & \rot{\emph{Forward}}  & \rot{\emph{Challenge}}   \\        
    \midrule
    \emph{Right} 
    & $\%$ Success   	& \textbf{99.0}  & 98.5  & 81.1  & 98.1  & 74.1  \\
    & $\%$ Collision 	& \textbf{0.96}  & 1.50  & 18.8  & 1.80  & 25.8  \\
    & Avg. Time 		& 4.70s & 4.51s & 4.39s & 4.60s & \textbf{4.02s} \\
    & Avg. Brake   		& 0.41s & 0.38s & \textbf{0.26s} & 0.42s & 0.43s \\
    \midrule
    \emph{Left} 
    & $\%$ Success   	& 87.7  & \textbf{96.3}  & 76.8  & 95.1  & 66.2  \\
    & $\%$ Collision    & 12.3  & \textbf{3.65}  & 23.1  & 4.84  & 33.7  \\
    & Avg. Time         & 5.35s & 5.36s & \textbf{5.31s} & 5.48s & 4.64s \\
    & Avg. Brake   		& 1.04s & 1.01s & \textbf{0.88s} & 1.01s & 1.04s \\
    \midrule
    \emph{Left2} 
    & $\%$ Success	 	& 70.1  & 76.3  & \textbf{91.7}  & 74.5  & 74.6  \\
    & $\%$ Collision    & 29.8  & 23.6  & \textbf{8.15}  & 25.4  & 25.4  \\
    & Avg. Time         & 6.02s & 5.90s & 6.48s & 6.06s & \textbf{5.38s} \\
    & Avg. Brake   		& 1.47s & 1.43s & \textbf{1.38s} & 1.43s & 1.53s  \\
    \midrule
    \emph{Forward}    
    & $\%$ Success	 	& 87.2  & 97.2  & 79.0  & \textbf{97.6}  & 69.9  \\
    & $\%$ Collision    & 12.7  & 2.74  & 21.0  & \textbf{2.34}  & 30.0  \\
    & Avg. Time         & 4.79s & 4.80s & 4.71s & 4.88s & \textbf{4.05s} \\
    & Avg. Brake   		& 1.05s & 1.02s & \textbf{0.91s} & 1.03s & 1.08s \\
    \midrule
    \emph{Challenge}    
    & $\%$ Success	 	& 58.7  & 60.7  & 72.3  & 62.4  & \textbf{78.5}  \\
    & $\%$ Collision    & 41.2  & 39.2  & 27.6  & 37.4  & \textbf{21.4}  \\
    & Avg. Time         & 6.73s & \textbf{6.58s} & 7.55s & 7.22s & 6.83s \\
    & Avg. Brake   		& 2.81s & 2.79s & \textbf{2.78s} & \textbf{2.78s} & 2.79s \\
\bottomrule
\end{tabular}
\end{center}
\label{table:direct_copy}
\vspace{-20pt}
\end{table}

\begin{figure*}[thpb!]
    \centering
    \vspace{8pt}
    \begin{subfigure}[b]{2.2in}
    	\includegraphics[height=.75\textwidth]{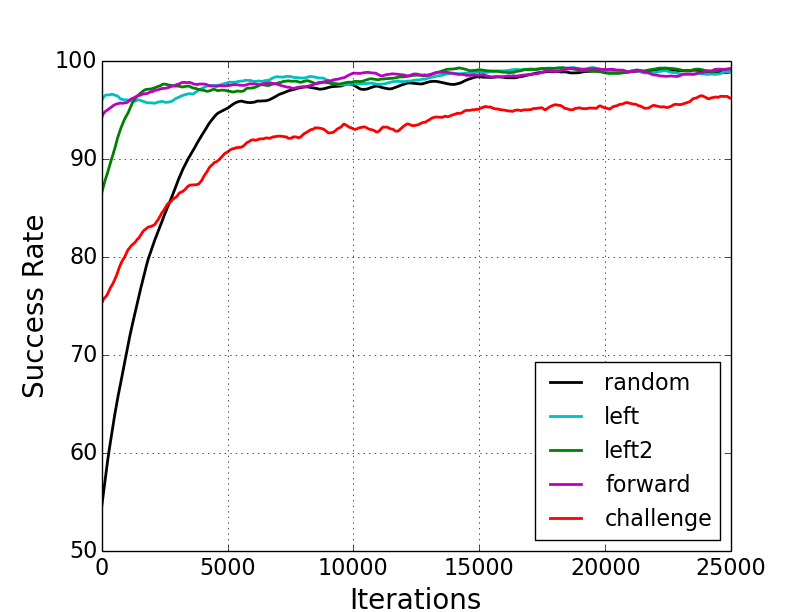} 
        \caption{\emph{Right}}
        \label{fig:scenarios_right}
    \end{subfigure}
    \begin{subfigure}[b]{2.2in}
    	\includegraphics[height=.75\textwidth]{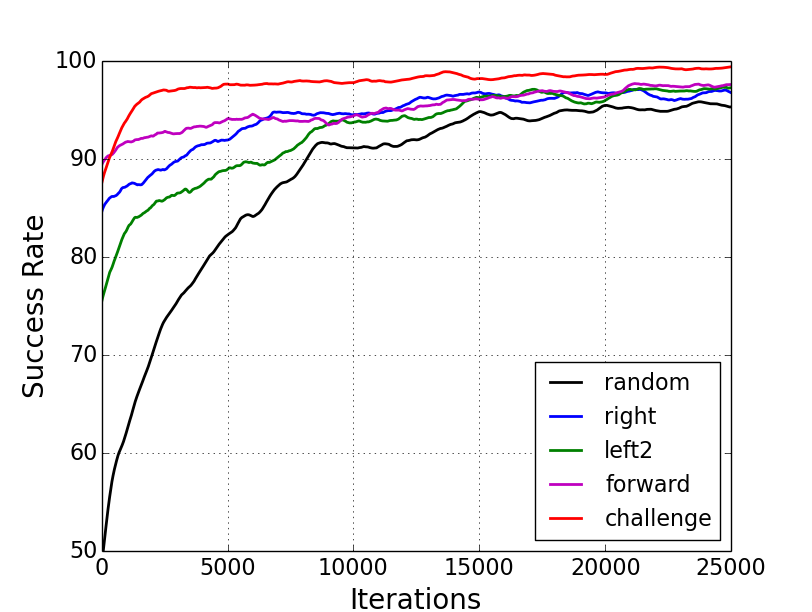}
        \caption{\emph{Left}}
        \label{fig:scenarios_left}
    \end{subfigure}
    \begin{subfigure}[b]{2.2in}
    	\includegraphics[height=.75\textwidth]{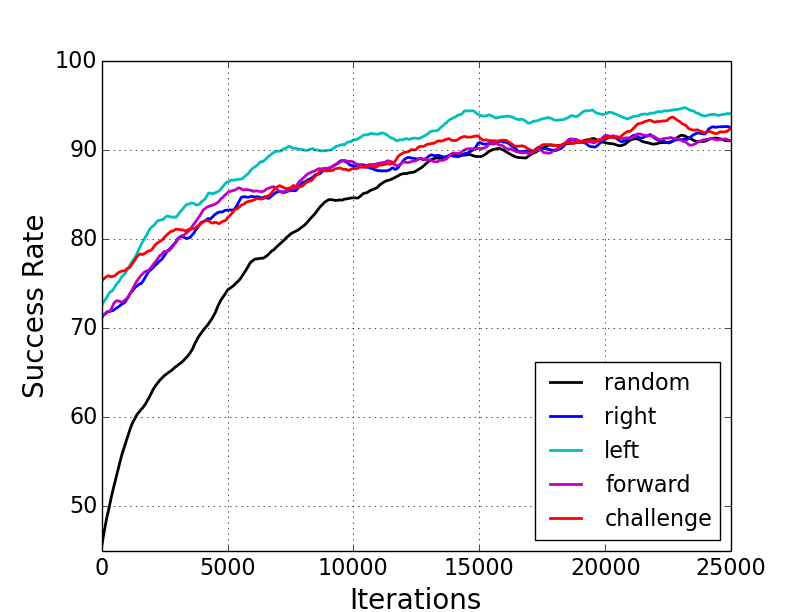}
        \caption{\emph{Left2}}
        \label{fig:scenarios_left2}
    \end{subfigure}
    \par\medskip
    \hspace{-4pt}
    \begin{subfigure}[b]{2.2in} 
       \includegraphics[height=.75\textwidth]{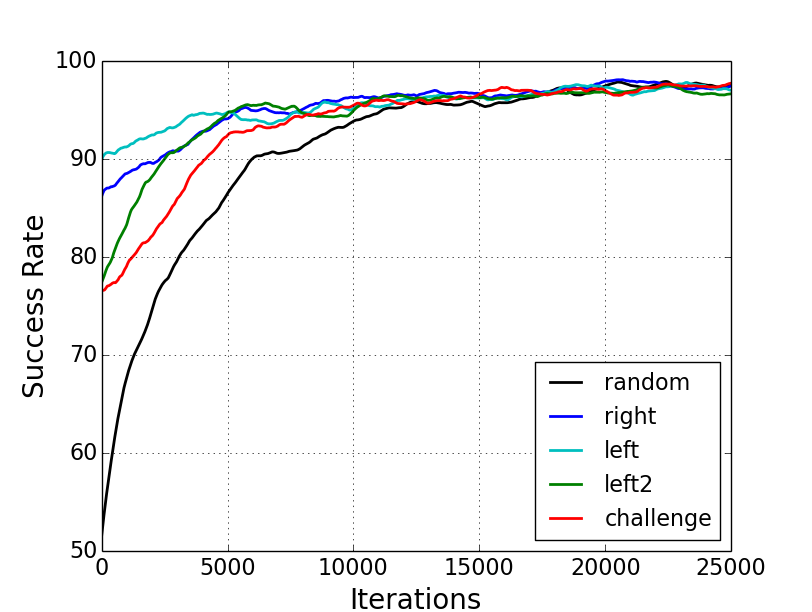}
        \caption{\emph{Forward}}
        \label{fig:scenarios_forward}
    \end{subfigure}
    \hspace{1pt}
    \begin{subfigure}[b]{2.2in}
    	\includegraphics[height=.75\textwidth]{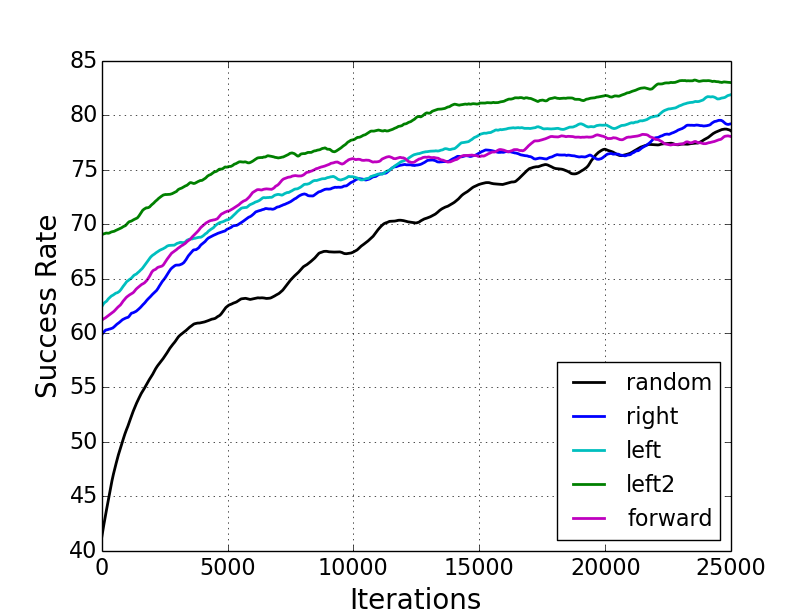}
        \caption{\emph{Challenge}}
        \label{fig:scenarios_challenge}
    \end{subfigure}
        \caption{Fine-tuning comparison. A network for one task is initialized with the network of a different task. The colored lines indicate the initialization network. The black line indicates the performance of a network trained with a random initialization. Initializing a network with a network trained on another task is almost always advantageous. We notice a jumpstart benefit in every tested example, and observe several asymptotic improvements. }\label{fig:fine-tune} 
        \vspace{-10pt}
\end{figure*}

\begin{figure}[t]
\centering
\hspace{-10pt}
  \includegraphics[width=.5\textwidth]{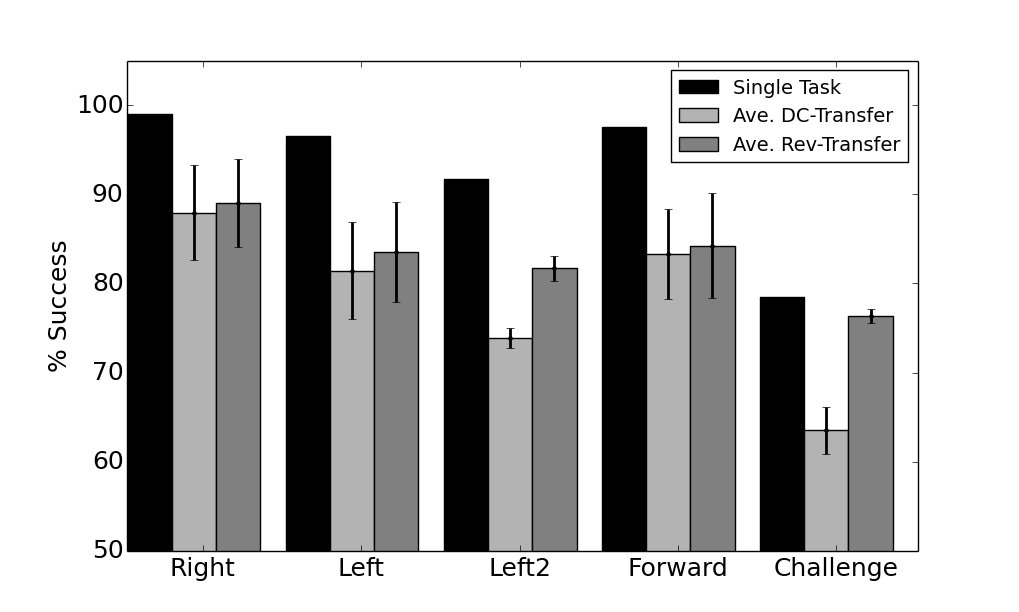}
  \caption{Direct Copy and Reverse Transfer. The x axis denotes the test condition. Black bars show the performance of single task learning. Light gray bars show the average performance of a network trained on one task and tested on another. The drop in performance demonstrates the difference between tasks. The dark gray indicates the average performance of reverse transfer: a network is trained on Task A, fine-tuned on Task B, and then evaluated on Task A. The drop in performance indicates catastrophic forgetting, but networks exhibit some retention of the initial task.}
  \label{fig:dc_rev}
  \vspace{-10pt}
\end{figure}

\begin{figure}[t]
\centering
\vspace{-5pt}
\hspace{-10pt}
  \includegraphics[width=0.5\textwidth]{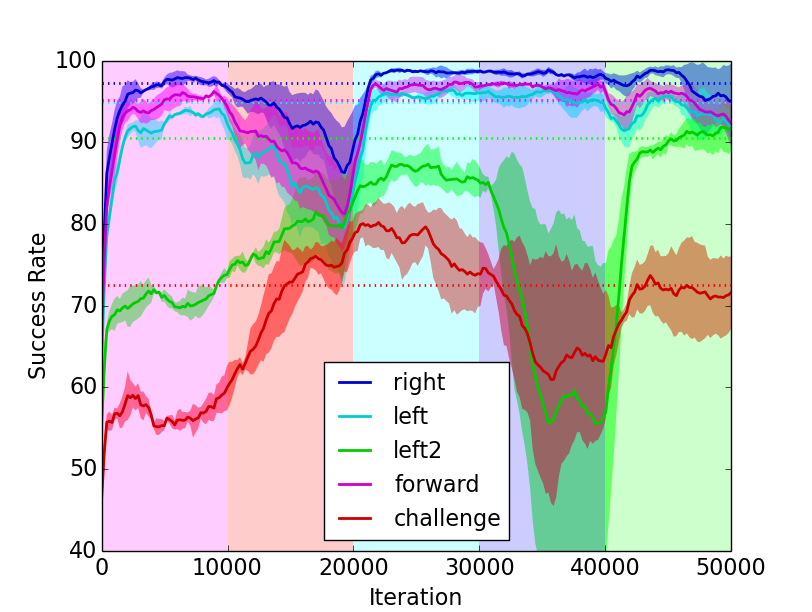}
  \caption{Lifelong learning results. The background color indicates which task is being trained. Each solid line depicts the test performance for a different task. Dotted lines indicate the performance of a network trained on a single task for an equivalent period of time. The standard deviation across runs is indicated by the envelope. At several points in the learning process, the network shows evidence of \emph{forgetting} previous tasks.}
  \label{fig:LL}
  \vspace{-10pt}
\end{figure}

\section{Results}
\label{sec:results}

\noindent\textbf{Direct Copy:} Table \ref{table:direct_copy} shows the results when a network is trained on one task and applied to another. In no instance does a network trained on a different task do better than a network trained on the matching task, but we do see that several tasks achieve similar performance with transfer. Particularly we see that the single lane tasks (right,left, and forward) are related, they are consistently the top performers in all single lane tasks. Additionally the more challenging multi-lane settings (left2 and challenge) appear connected, the \emph{Left2} network does substantially better than either of the single lane tasks on the \emph{Challenge} task.  

\noindent\textbf{Fine Tuning:} Figure \ref{fig:fine-tune} shows fine tuning results. We see that in nearly all cases, pre-training with a different network gives a significant advantage in \emph{jumpstart} \cite{taylor2009transfer} and in several cases there is an asymptotic benefit as well. When the fine tuned networks are re-applied to the source task the performance looks similar to direct copy, as shown in Figure \ref{fig:dc_rev}. 

\noindent\textbf{Reverse Transfer:} While the performance on the source task drops after fine tuning, we see a trend of positive improvement compared to direct copy. This  indicates that some information was retained by the network. Figure \ref{fig:retention} shows the retention for each task pair, showing the percentage gain resulting from the initialization. The \emph{Left2} and \emph{Challenge} tasks have less overlap with other tasks in the state space, so it is possible that more aspects of the initialization are left unchanged, which might explain why there is the largest amount of retention for these tasks. This hypothesis is supported by the fact that training on the \emph{Right} task exhibits the most retention, since these two tasks have the least overlap.  

\noindent\textbf{Lifelong learning:} The results for the lifelong learning experiment are shown in Figure \ref{fig:LL}. Every task initially benefits from learning on the first task (\emph{Forward}), although the performance in the \emph{Left2} and \emph{Challenge} settings benefit less. In some cases we see that training on a different task helps up to a point and then further training hurts other tasks. For example, after training on approximately 5000 trials of \emph{Forward} setting, the \emph{Right} task performance starts to decrease. 

Overall, we see an affinity between both the single lane tasks (\emph{Left}, \emph{Right}, and \emph{Forward}) and the multi-lane tasks. When training on the \emph{Challenge} task starts, \emph{Left2} benefits, but the single lane tasks exhibit catastrophic forgetting. Training on the \emph{Left} task helps the other single lane tasks, but \emph{Challenge} decreases in performance. 

However the results are not consistent across the grouping of single lane tasks. Training on the \emph{Right} task has a much more detrimental effect on the multi-lane tasks than either \emph{Forward} or \emph{Left}. We suspect this is because right turns can ignore one of the lanes of traffic which matters to all other tasks. Overall the negative effects of catastrophic forgetting negate many of the positive effects of transfer. 

\section{Conclusion}
\label{sec:conclusion}

In this paper we view the AD vehicle as a learning agent in a reinforcement learning setting, and analyze how the knowledge for handling one type of intersection, represented as a Deep Q-Network, translates to other types of intersections. We investigated and compared four different transfer methods between different intersections (tasks): direct copy, fine tuning, reverse transfer and lifelong learning. Our results have several conclusions. First, we found the success rates were consistently low when a network is trained on Task A but directly tested on Task B. Second, a network that is initialized with the network of a Task A and then fine-tuned on Task B generally performed better than a randomly initialized network that is trained on Task B. Third, when a network that is initialized with Task A, fine-tuned on Task B, and is tested back on Task A, it performed better than a network directly copied from Task B to Task A. Finally, we examine a lifelong learning domain, where we train a single network to handle all five intersection scenarios and show that the resulting network exhibited catastrophic forgetting of previous task knowledge.

As future work, we will conduct research on the concept of a long-term memory and investigate how to effectively preserve previous task knowledge for lifelong learning.


\bibliographystyle{IEEEtran}
\bibliography{refs}

\end{document}